\pdfoutput=1

\documentclass[11pt]{article}

\usepackage{latex/acl}

\usepackage{times}
\usepackage{latexsym}

\usepackage[T1]{fontenc}

\usepackage[utf8]{inputenc}

\usepackage{microtype}

\usepackage{inconsolata}
\usepackage{graphicx} 
\usepackage[namelimits]{amsmath} 
\usepackage{amssymb}
\usepackage{booktabs} 
\usepackage{multirow}
\usepackage{makecell}
\usepackage{pgfplots}
\usepackage{color}

\title{Hypergraph based Understanding for Document Semantic Entity Recognition}

\author{Qiwei Li$^{1}$\footnotemark[2], Zuchao Li$^{1}$\footnotemark[1]\footnotemark[2], Ping Wang$^{2,4}$, Haojun Ai$^3$\footnotemark[1] and Hai Zhao$^5$\\
{ $^{1}$School of Computer Science, Wuhan University} \\
{ $^{2}$School of Information Management, Wuhan University} \\
{ $^{3}$School of Cyber Science and Engineering, Wuhan University} \\
{ $^{4}$Key Laboratory of Archival Intelligent Development and Service, NAAC} \\
{ $^{5}$Department of Computer Science and Engineering, Shanghai Jiao Tong University}\\
{\tt \{qw-line,zcli-charlie,wangping,aihj\}@whu.edu.cn}\\
}

\begin{document}

\maketitle

\renewcommand{\thefootnote}{\fnsymbol{footnote}} 
\footnotetext[1]{Corresponding author.} 
\footnotetext[2]{Equal contribution. }
\renewcommand{\thefootnote}{}
\footnotetext{This work was supported by the National Natural Science Foundation of China (No. 62306216, No. 72074171, No. 72374161), the Natural Science Foundation of Hubei Province of China (No. 2023AFB816), the Fundamental Research Funds for the Central Universities (No. 2042023kf0133). }

\begin{abstract}
Semantic entity recognition is an important task in the field of visually-rich document understanding. It distinguishes the semantic types of text by analyzing the position relationship between text nodes and the relation between text content. The existing document understanding models mainly focus on entity categories while ignoring the extraction of entity boundaries. We build a novel hypergraph attention document semantic entity recognition framework, HGA, which uses hypergraph attention to focus on entity boundaries and entity categories at the same time. It can conduct a more detailed analysis of the document text representation analyzed by the upstream model and achieves a better performance of semantic information. We apply this method on the basis of GraphLayoutLM to construct a new semantic entity recognition model HGALayoutLM. Our experiment results on FUNSD, CORD, XFUND and SROIE show that our method can effectively improve the performance of semantic entity recognition tasks based on the original model. The results of HGALayoutLM on FUNSD and XFUND reach the new state-of-the-art results.
\end{abstract}

\section{Introduction}

With the development of information technology, documents have become a main information carrier nowadays, which contains kinds of information type, such as text, table and image. Manual recognition of these documents often requires plenty of manpower. OCR tools can only help us to identify the text, layout and other simple information in the document. To further understand documents, Visually-rich Document Understanding (VRDU)~\cite{xu2020layoutlm} is proposed to make use of visual, textual and other information for more in-depth analysis.

\begin{figure}[t]
\centering
\includegraphics[width=0.45\textwidth]{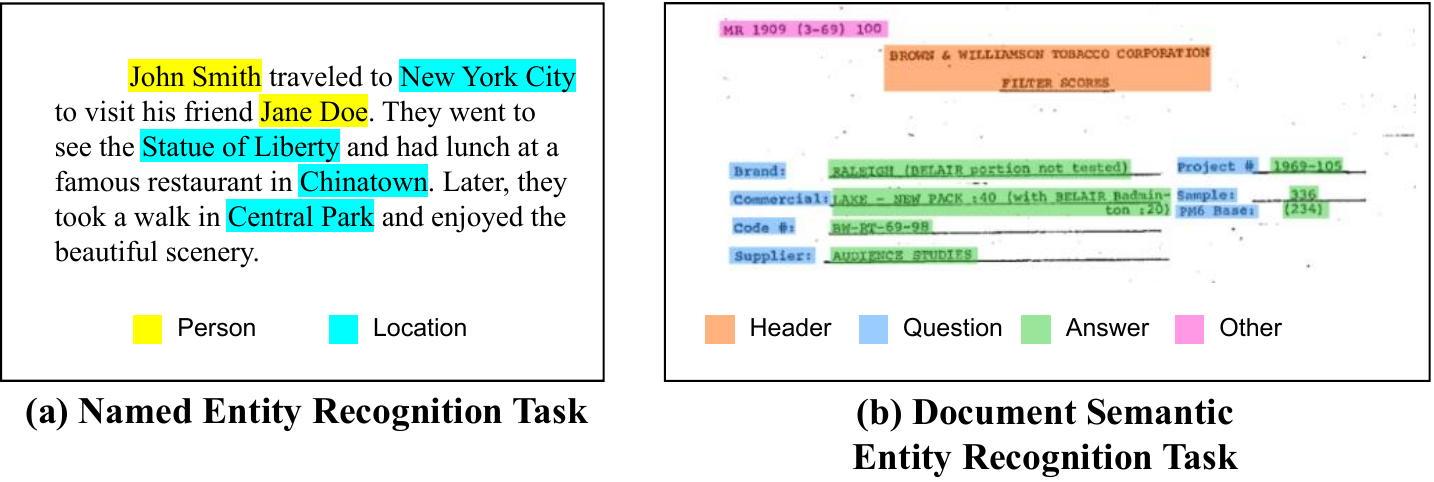}
\caption{Difference in Document Task.}
\label{fig:difference_in_document}
\end{figure}

Semantic Entity Recognition (SER) is an important task in the field of VRDU. Its purpose is to extract and classify the text with special semantic information in documents. Different from text sequences in traditional natural language processing tasks, the information in documents is not one-dimensional, single-modal and continuous, but two-dimensional, multimodal and discrete. It is necessary to analyze not only text information, but also other modal information such as layout and vision in the document. Figure \ref{fig:difference_in_document} shows the difference between the traditional named entity recognition (NER) task on a single modal text and the semantic entity recognition task on a document. Firstly, the text form of a single modal text task is a fixed text sequence, while the discrete text in a document is composed of text nodes in different locations. Secondly, the named entity recognition task of a single modal text only needs to consider the semantic relationship between the tokens in the text sequence. However, the semantic entity recognition task on the document needs to consider not only the semantic relationship between nodes, but also the position relationship between nodes. Finally, the span range of entity tags of NER task is flexible, while the range of task tags of semantic entity recognition task on document is affected by nodes. Texts of the same node in the document share the same label in most cases.

With the development of pre-training technology, document pre-training model has become popular. LayoutLM ~\cite{xu2020layoutlm} is the first multi-modal pre-trained model to associate text with layout and vision, achieving leading results on multiple downstream document understanding tasks including semantic entity recognition. Subsequently, more multi-mode pretraining models, such as LayoutLMv2~\cite{xu2020layoutlmv2}, BROS~\cite{hong2022bros}, ERNIE-Layout~\cite{peng2022ernie} and LayoutLMv3~\cite{huang2022layoutlmv3} have been proposed successively. By integrating text, layout and visual information, they realize the understanding and information extraction of documents. So far, GraphLayoutLM~\cite{li2023enhancing} and GeoLayoutLM~\cite{luo2023geolayoutlm} have the best performance in semantic entity recognition tasks. GraphLayoutLM achieves the best F1 score of 94.39 and 93.56 on the FUNSD~\cite{jaume2019funsd} and XFUND~\cite{xu2021layoutxlm} datasets.  GeoLayoutLM achieves the best F1 score of 97.97 on the CORD~\cite{park2019cord} datasets. However, these existing methods focus on the upstream document understanding part and pay little attention to the downstream task. GeoLayoutLM has studied the novel relational extraction head and achieves great improvement in the relational extraction task. But it has not done more research on the semantic entity recognition task. We study the problem of ignoring the downstream head and classification method in the semantic entity recognition task in the existing document intelligence work and propose a novel improvement scheme.

\begin{figure*}[ht]
\centering
\includegraphics[width=0.95\textwidth]{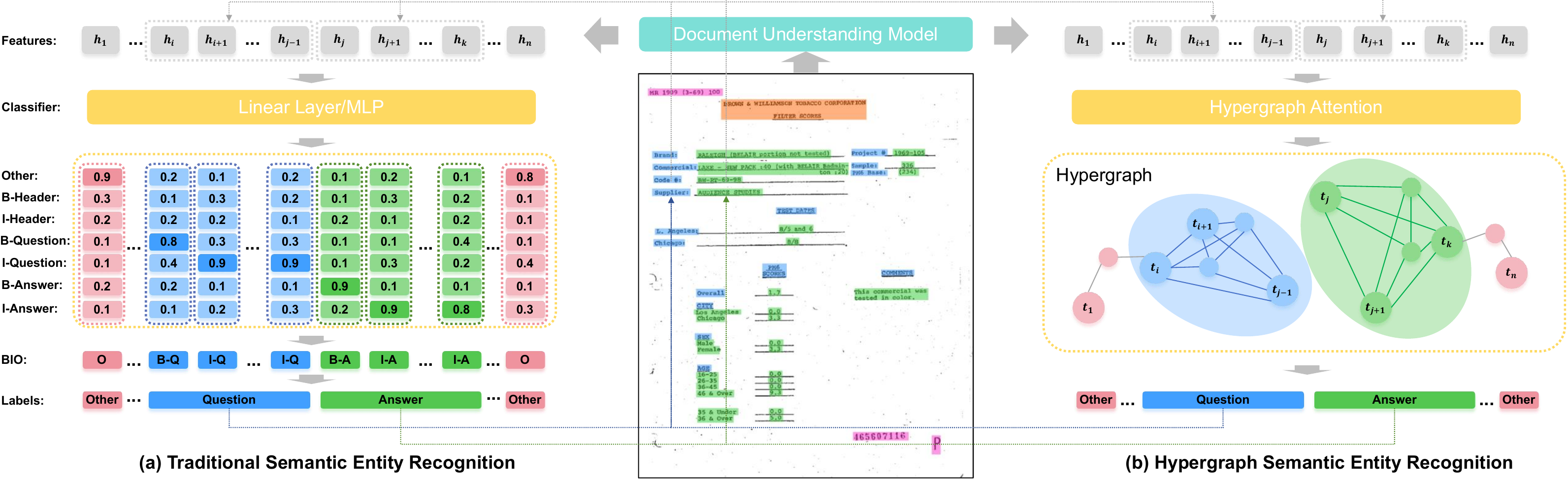}
\caption{Traditional Semantic Entity Recognition and Hypergraph Semantic Entity Recognition.The document is from FUNSD dataset. Only the text sequence is shown in the figure. The rectangles with different colors in the figure are text nodes. The colors on the document nodes represent the different class labels. The orange color represents the label "HEADER". Blue is the label "QUESTION". Green is the label "ANSWER". Pink is the nonmeaning label, which is "OTHER".}
\label{fig:different_ser}
\end{figure*}

\paragraph{Traditional Semantic Entity Recognition.} The traditional document semantic entity recognition task process is shown in (a) of the Figure \ref{fig:different_ser}. In document understanding process, text nodes are spliced into text sequences and become text token sequences of documents after tokenization. These text nodes will be transformed to the high-dimensional feature representations after the analysis of the document understanding model. To extract semantic information from document token features, linear layer or multilayer perceptron (MLP) will be used to convert high-dimensional features into label probabilities and the training objective is cross entropy loss. Although this method can distinguish the node categories in the document, it ignores the characteristics of the document structure and it is difficult to make the classification layer pay attention to the node span.

\paragraph{Hypergraph Semantic Entity Recognition.} Inspired by Global Pointer~\cite{su2022global}, we use the idea of hypergraph to extract the semantic information of documents and propose a Hypergraph Attention(HGA) strategy for document semantic entity recognition. (b) of the Figure \ref{fig:different_ser} shows us the process of  hypergraph semantic recognition. Different from the traditional classification method, the semantic entity recognition idea of HGA regard the document token features as graph nodes. The target entity is the set of nodes with the same hyperedge and the hyperedge type represents the entity label type. The process of hypergraph extraction is to establish hyperedges between token feature nodes. Besides, we use the span hyperedge encoding to add the span information of text nodes. Through the hypergraph and span position, the head can better focus on the entity boundary information and establish the relationship between the document discrete text span and the entity boundary.

Our main contributions are as follows:

\begin{itemize}
\item We construct a novel hypergraph attention document semantic entity recognition method, HGA. It transforms the traditional token sequence classification problem into a hypergraph construction process. By establishing different types of hyperedges between text nodes, the head can extract semantic entities.

\item We propose a novel span hyperedge position encoding and  balanced hyperedge loss.  Span hyperedge position encoding makes the head focus more on the same text span prompt during hyperedge construction. Balanced hyperedge loss can help to solve the problem of matrix sparsity caused by too many hyperedge types in some scenarios.

\item We construct a novel document semantic entity recognition model HGALayoutLM based on the HGA method. Our code will be available at
https://github.com/Line-Kite/HGALayoutLM. The experiment results show that the model has good performance in the scene with few types of labels. HGALayoutLM has obtained the best results on the FUNSD, SROIE and XFUND datasets. 
\end{itemize}

\section{Related Work}

In recent years, self-supervised pre-training technology has become the mainstream trend in the fields of natural language processing (NLP) and computer vision (CV).  BERT~\cite{devlin2018bert} is a classic pre-training model that has shown great effectiveness in various tasks such as question answering, natural language generation and text classification. Masked Language Modeling (MLM) is a significant pre-training task proposed by BERT that enables models to learn textual representations by predicting the raw vocabulary ids of randomly masked word markers based on context. Since then, a series of mask language models such as RoBERTa~\cite{liu2019roberta}, ALBERT~\cite{lan2019albert} and XLNet~\cite{yang2019xlnet} have been proposed successively. These models achieve good results on natural language understanding tasks.

However, the single modal language model~\cite{lan2019albert,liu2019roberta,peng2023fsuie,li2023batgpt} can not understand documents with complex formats and diverse types well. To fully understand the content of complex documents,  LayoutLM~\cite{xu2020layoutlm} adds layout and document information on the basis of BERT to supplement the document format missing from plain text. Following LayoutLM, BROS~\cite{hong2022bros}, LayoutLMv2~\cite{xu2020layoutlmv2}, XYLayoutLM~\cite{gu2022xylayoutlm}, ERNIE-Layout~\cite{peng2022ernie}, LayoutLMv3~\cite{huang2022layoutlmv3} and other multi-modal pre-training document understanding models have been proposed successively and constantly make breakthroughs in various tasks in the field of document understanding. These models understand the document through the fusion of text, layout and vision information. Since document nodes are suitable to be represented by graph structures, some works begin to apply graph structures to document understanding models, such as ERNIE-mmLayout~\cite{wang2022ernie}, ROPE~\cite{lee2021rope}, FormNet~\cite{lee2022formnet} and GraphLayoutLM~\cite{li2023enhancing}.

\begin{figure*}[t]
\centering
\includegraphics[width=0.9\textwidth]{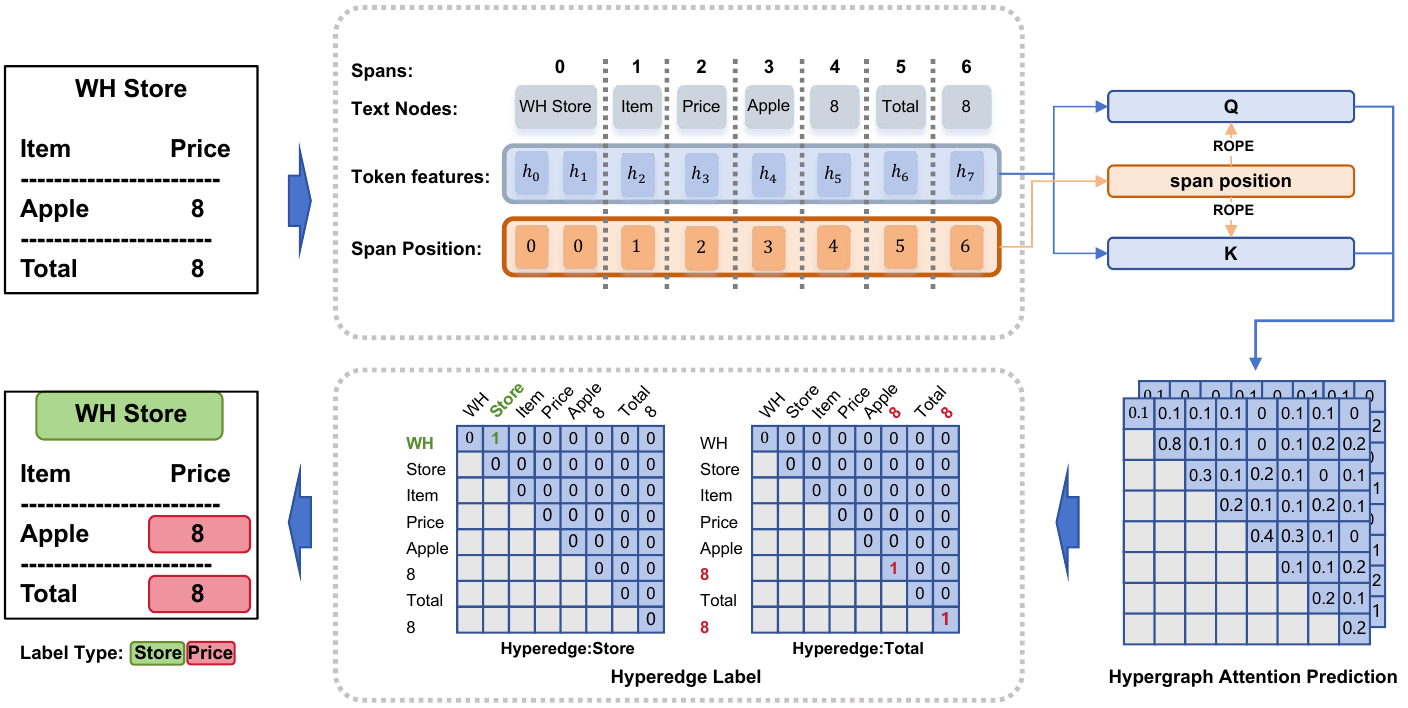}
\caption{Semantic Entity Recognition Process Based on Hypergraph Attention. Only the text processing part of the model is shown in the figure. In the span position generation stage, the span position of the token feature sequence needs to be created by using the text node range span.  The token features will be linearly transformed and encode the span position into a query vector Q and a key vector V. The multi-head hypergraph attention score is calculated from Q, V and added with the lower triangle mask.  We regard each attention head as a sub-hypergraph corresponding to each hyperedge type.}
\label{fig:hga}
\end{figure*}

The latest GraphLayoutLM and GeoLayoutLM~\cite{luo2023geolayoutlm} are both built on the basis of LayoutLMv3. They have achieved the most excellent results in several tasks of document information extraction. Inspired by some graph based works~\cite{kipf2016semi,velickovic2017graph,wang2023enhancing}, GraphLayoutLM models the document structure based on the hierarchical and positional layout of the document and represents the document layout modeling with a graph structure. To integrate graph structure information into the process of document understanding, GraphLayoutLM proposes graph reordering and graph masking strategies, adding graph information into the document understanding model in the form of sequence and self-attention mask. GeoLayoutLM implements geometric pre-training to enrich and enhance feature representation through three specially designed geometry-related pre-training tasks. In addition, GeoLayoutLM uses a novel relation head in the fine-tuning phase and obtains a big improvement over LayoutLMv3 in the relation extraction task. At present, little attention is paid to the effects of downstream task heads on the performance of various types of tasks. GeoLayoutLM proposes a novel relational head, but there is still a lack of research on the downstream task of semantic entity recognition in the field of document understanding. Most of the current models use a linear layer and cross-entropy to predict BIO label probabilities when dealing with semantic entity recognition tasks, such as LayoutLM, BROS, LayoutLMv2, etc. LayoutLMv3 and its derived models utilize a linear layer in the few label case and employ MLP when number of label types is large. These approaches are fundamentally the same.  Differently, UDop~\cite{tang2023unifying} is a new unified document intelligent framework, which adopts encoder-decoder structure. In addition, with the development of large language models (LLMs), some works~\cite{hong2023cogagent,fujitake2024layoutllm,luo2024layoutllm} have begun to apply large model technology to document intelligence. However, the decoder will cost a large computational cost. Taking inspiration from Global Pointer~\cite{su2022global}, we design a simple hypergraph head that incorporates document span information to achieve better SER task performance.

\section{Methodology}

\subsection{Overview}

The process of semantic entity recognition based on Hypergraph Attention is shown in Figure \ref{fig:hga}. Different from traditional semantic entity recognition methods, HGA focuses on extracting special entities. Instead of using BIO annotation method as the label annotation strategy for model input, we regard each special label (such as header, question and answer) as a label type. The entity without special meaning represented by the Other label in the BIO annotation will not be labeled as a hyperedge type in the hypergraph construction process. HGA regards token features as unit nodes and the process of establishing hyperedges between tokens can realize the extraction of special entities. It is worth noting that the node referred to here correspond to each token of token sequence. Text nodes, as mentioned earlier, are discrete pieces of text at different locations in the document. A text node corresponds to one or more token feature nodes. The process of hyperedge extraction can realize the extraction of special semantic entites and classification of different entity labels.
Entities that do not have the meaning of a special label (that is, the entities of the Other label in the BIO annotation) will not have the connection of any special hyperedge.

To assist the construction of hyperedges, we use the span of each text node to generate the span position corresponding to the feature sequence. Then we use the span position encoding to add span information to the hypergraph construction process.   In this way, the model can divide the hyperedge according to the text node span, so as to achieve more accurate extraction of the special entity range.   In the stage of semantic entity extraction, we use multi-label classification to determine whether a node is connected by a hyperedge.  Since there may be more than one type of hyperedges satisfying the join condition.  To ensure the uniqueness of the entity type, we select the hyperedge with the maximum probability to establish the connection based on multi-label classification result.

\subsection{Hypergraph Attention Head}

We use the multi-head self-attention matrices to represent the hypergraph. Consider a hypergraph with $L$ number of nodes and $N$ class of hyperedges. We use a multihead attention score of shape $N \times L \times L$ as the representation of this hypergraph. Hyperedge classes are represented by different heads of multi-head attention. The attention matrix corresponding to each head represents the distribution of a type hyperedge.

In the hypergraph, each token corresponds to a node. Assume the document token sequence is $x=\{x_1,x_2,...,x_n\}$. After understanding the document model, we convert the input token sequence into a high-dimensional feature representation sequence of the tokens:
\begin{equation}
h=\{h_1,h_2,...,h_n\}=\text{Model}(\{x_1,x_2,...,x_n\}),
\end{equation}
where $h\in \mathbb{R}^{L\times H}$ is the high-dimensional feature representation sequence of the token and $\text{Model}(\cdot)$ is the document understanding model. $L$ indicates the token sequence length, which also represents the number of token nodes. $H$ is the feature dimension size. Based on $h$, we can obtain the query vector $q$ and the key vector $k$:
\begin{equation}
\begin{aligned}
q=\{q_{\alpha} : W_{q,\alpha}h+b_{q,\alpha}\},\\
k=\{k_{\alpha} : W_{k,\alpha}h+b_{k,\alpha}\},
\end{aligned}
\end{equation}
where $\alpha \in \mathbb{Z}^{D}$ is one head in multi-head attention, which can be regarded as a type in $D$ kinds of hyperedges. With multi-head query vector and key vector, hypergraphs can be represented by a self-attention score calculated by $q$ and $k$:
\begin{equation}
s=q^Tk=\{s_{\alpha}(i,j):q_{i,\alpha}^Tk_{j,\alpha}, i \in \mathbb{Z}^{L}, j \in \mathbb{Z}^{L}\}.
\end{equation}
$s_{\alpha}(i,j)$ is the attention score at the $\alpha$ type hyperedge span with $[i,j]$. $q_{i,\alpha}$ and $k_{j,\alpha}$ are the start and end of the span with $[i,j]$ in the $\alpha$ type hyperedge matrix. In this way, we implement hypergraph extraction of semantic entities.

\subsection{Span Position Encoding}

As we mentioned in Introduction,  tokens of the same text node normally share the same semantic label in the process of semantic entity recognition of documents.  We hope that the head can consider this span boundary prompt during entity extraction. Therefore, we construct the span position of the token sequence based on the text nodes and incorporate span information into the heads through position encoding. As shown in Figure \ref{fig:hga}, token feature sequence $h\{h_1,h_2,...,h_n\}$ and text node sequence $N=\{N_0,N_1,...,N_m\}$ has a surjective relation. We define this relational mapping as:
\begin{equation}
f(h_i)=N_j,h_i \in h, N_j \in N.
\end{equation}
Based on this relation mapping, we construct the span position. For the same text node $N_j$, all token feature nodes that have a mapping relationship with the same text node $N_j$ share the same position:
\begin{equation}
\begin{aligned}
p_i &=Position(f(h_i))  \\
& =Position(N_j) \\
&=j,h_i \in h, N_j \in N,
\end{aligned}
\end{equation}
where $p_i$ is the span position of token feature $h_i$, $Position$ is the index of $N_j$. In this way, we can obtain the span position sequence $p=\{p_1,p_2,...,p_n\}$. On the basis of $p$, we use rotary position coding~\cite{su2021roformer} to generate position encoding $\mathcal{R}$, which satisfies $\mathcal{R}^T_i \mathcal{R}_j = \mathcal{R}_{j-i}$. Then the calulation of multi-head hypergraph score will be adjust to the following form:
\begin{equation}
\begin{aligned}
s_{\alpha}(i,j)&=(\mathcal{R}_ip_{i,\alpha})^T(\mathcal{R}_jk_{j,\alpha}) \\
&=p_{i,\alpha}^T\mathcal{R}_i^T\mathcal{R}_jk_{j,\alpha} \\
&=p_{i,\alpha}^T\mathcal{R}_{j-i}k_{j,\alpha}.
\end{aligned}
\end{equation}
Because the start is always before the end when the span of token sequence is extracted.  Span extraction nodes should not appear in the lower triangular region of the hypergraph attention score.  For the purpose of making the hyperedge construction more reasonable, we add $m_{tril}$ to the hypergraph matrix and the final hypergraph score format is as follow:
\begin{equation}
s_{\alpha}(i,j)=p_{i,\alpha}^T\mathcal{R}_{j-i}k_{j,\alpha}+m_{tril}(i,j).
\end{equation}

\subsection{Balanced Hyperedge Loss}

In the process of loss calculation, we collect positive samples $P_{\alpha}$ and negative samples $N_{\alpha}$ respectively for each type of hyperedge $\alpha$ . The positive sample indicates that there is a $\alpha$ type hyperedge span with $[i,j]$ in $\alpha$ type hypergraph, while the reverse is a negative sample. The formats of $P_{\alpha}$ and $N_{\alpha}$ are as follows:
\begin{equation}
\begin{aligned}
P_{\alpha}=\{s_{\alpha}(i,j)|l_{\alpha}(i,j)=1\},\\
N_{\alpha}=\{s_{\alpha}(i,j)|l_{\alpha}(i,j)=0\},
\end{aligned}
\end{equation}
where $l$ is the hypergraph label matrix corresponding to $s$. With the sets of positive and negative samples,we can get the positive sample loss $\mathcal{L}_{p}$ and the negative sample loss $\mathcal{L}_{n}$:
\begin{equation}
\begin{aligned}
\mathcal{L}_{p}=\log \left(1 + \sum\limits_{(i,j)\in P_{\alpha}} e^{-s_{\alpha}(i,j)}\right),\\
\mathcal{L}_{n}=\log \left(1 + \sum\limits_{(i,j)\in N_{\alpha}} e^{s_{\alpha}(i,j)}\right).
\end{aligned}
\end{equation}
Different from Global Pointer~\cite{su2022global}, we gain the final loss with a balance factor $b \in [0,1)$ to avoid the matrix sparsity caused by too many label types. The final training loss of hypergraph attention score can be expressed in the following form:
\begin{equation}	
\mathcal{L}=(1+b)\mathcal{L}_{p} + (1-b)\mathcal{L}_{n}.
\end{equation}

\subsection{HGALayoutLM}

To verify the performance of the HGA method, we apply HGA to the latest GraphLayoutLM to build a novel semantic entity recognition model, HGALayoutLM.  We use GraphLayoutLM as the base model for feature encoding. According to its input requirements, we input four multi-modal document information of the document: text, layout, visual and graph to obtain the feature sequence of the text tokens. Before input, we sort the sequence of text tokens using the layout graph according to the reordering strategy of GraphLayoutLM. On the basis of this graph structure-prompted document understanding model, we use the hypergraph attention layer as the head for document semantic entity recognition. The feature sequence of the token and the generated span position are used as the head input. The HGA method is used to help the model extract and classify semantic entities according to the text node span prompts.

\section{Experiment}

\subsection{Experimental Setup}

\begin{table}[t]
    \centering
    \caption{Detail Data of Datasets. The nonmeaning label "OTHER" is not included.}
    \label{tab:datasets_data}
    \begin{tabular}{lcccc}
    \toprule
     \bf Dataset & \bf Label Num & \bf Train & \bf Dev & \bf Test \\
     \midrule 
     FUNSD &3&149&-&50\\
     CORD &30&800&100&100\\
     SROIE &4&626&-&347\\
     XFUND &3&149&-&50\\
     \bottomrule
    \end{tabular}
\end{table}

\begin{table}[t]
    \centering
    \small
    \caption{Finetuning Hyper-parameters. L, M, B and G refer to learning rate, max steps, batch size and gradient accumulation steps. }
    \label{tab:hyper_parameter}
    \begin{tabular}{c|c|c|cccc}
        \toprule
        \bf Dataset & \bf \makecell[c]{Model \\ size} & \bf Language &\bf L & \bf M & \bf B & \bf G \\
        \midrule
        \multirow{2}{*}{FUNSD}
        & BASE & \multirow{2}{*}{English} & 1e-5 & 2000 & 4 & 1\\
        \multirow{2}{*}{}
        & LARGE & & 1e-5 & 2000 & 4 & 1\\
        \toprule
        \multirow{2}{*}{CORD}
        & BASE & \multirow{2}{*}{English} & 5e-5 & 2000 & 4 & 1\\
        \multirow{2}{*}{}
        & LARGE & & 5e-5 & 3000 & 4 & 1\\
        \toprule
        \multirow{2}{*}{SROIE}
        & BASE & \multirow{2}{*}{English} & 1e-5 & 2000 & 4 & 1\\
        \multirow{2}{*}{}
        & LARGE & & 1e-5 & 2000 & 4 & 1\\
        \toprule
        XFUND & BASE & CHINESE & 7e-5 & 2000 & 8 & 4 \\
        \bottomrule
    \end{tabular}
\end{table}

\begin{table*}[t]
    \centering
    \caption{Precision, Recall and F1 Score of Results on FUNSD, CORD, SROIE Datasets. Model labeled with "$^\dagger$" indicate that its results are obtained through replication in our experiments.
    The grey score of LayoutLMv3 on the SROIE dataset indicates that some of LayoutLMv3's predictions on the web based on the SROIE dataset were completely correct and we did not successfully reproduce its results. So we do not use it as a comparison.}
    \label{tab:main_results}
    \scalebox{0.85}{
    \begin{tabular}{l|l|ccc|ccc|ccc}
    \toprule
    \multirow{2}{*}{\bf Model} & \multirow{2}{*}{\bf Head} & \multicolumn{3}{c|}{\bf FUNSD}  & \multicolumn{3}{c|}{\bf CORD}  & \multicolumn{3}{c}{\bf SROIE} \\
    & & \bf P & \bf R  & \bf F & \bf P & \bf R  & \bf F & \bf P & \bf R  & \bf F \\
     \midrule 
     $\textrm{BERT}_{\rm BASE}$ & Linear & 54.69 & 67.10 & 60.26 & 88.33 & 91.07 & 89.68 & 90.99 & 90.99 & 90.99 \\
     $\textrm{LayoutLM}_{\rm BASE}$ & Linear & 75.97 & 81.55 & 78.66 & 94.37 & 95.08 & 94.72 & 94.38 & 94.38 & 94.38 \\
     $\textrm{BROS}_{\rm BASE}$ & Linear & 81.16 & 85.01 & 83.05 & - & - & 96.50 & - & - & 96.28 \\
     $\textrm{LayoutLMv2}_{\rm BASE}$ & Linear & 80.29 & 85.39 & 82.76 & 94.53 & 95.39 & 94.95 & 96.25 & 96.25 & 96.25\\
     $\textrm{LayoutXLM}_{\rm BASE}$ & Linear & - & - & 79.40 & -  & - & - & -  & - & -  \\
     $\textrm{XYLayoutLM}$ & Linear & - & - & 83.35 & - & - & - & - & - & - \\
     $\textrm{LayoutLMv3}_{\rm BASE}$ & Linear/MLP & 90.82 & 91.55 & 91.19 & 96.35 & 96.71 & 96.53 & {\color{gray} 100} & {\color{gray} 100} & {\color{gray} 100} \\
     $\textrm{GraphLayoutLM}_{\rm BASE}$ & Linear/MLP & 92.46 & \bf 93.85 & 93.15 & 97.02 & \bf 97.53 &  97.28 & - & - & 99.30  \\
     \midrule 
      $\textrm{GraphLayoutLM}_{\rm BASE}^\dagger$ & Linear/MLP & 93.62 & 93.25 & 93.43 & 96.87 & 97.38 &  97.13 & 98.40 & \bf 99.58 & 98.99 \\
     $\textrm{HGALayoutLM}_{\rm BASE}$ & HGA & \bf 94.84 &  93.80 & \bf 94.32 & \bf 97.89 & 97.16 & \bf 97.52 & \bf 99.58  & 99.48  & \bf 99.53  \\
     \midrule 
     $\textrm{BERT}_{\rm LARGE}$ & Linear & 61.13 & 70.85 & 65.63 & 88.86 & 91.68 & 90.25 & 92.00 & 92.00 & 92.00 \\
     $\textrm{LayoutLM}_{\rm LARGE}$ & Linear & 75.69 & 82.19 & 78.95 & 94.32 & 95.54 & 94.93 & 95.24 & 95.24 & 95.24 \\
     $\textrm{BROS}_{\rm LARGE}$ & Linear & 82.81 & 86.31 & 84.52 & - & - & 97.28 & - & - & 96.62 \\
     $\textrm{LayoutLMv2}_{\rm LARGE}$ & Linear & 83.24 & 85.19 & 84.20 & 95.65 & 96.37 & 96.01 & 99.04 & 96.61 & 97.81 \\
     $\textrm{ERNIE-Layout}_{\rm LARGE}$ & Linear & - & - & 93.12 & - & - & 97.21 & - & - & 97.55 \\
     $\textrm{LayoutLMv3}_{\rm LARGE}$ & Linear/MLP & 91.51 & 92.70 & 92.10 & 97.45 & 97.52 & 97.49 & - & - & - \\
     $\textrm{UDop}$ & Decoder & - & - & 92.08 & - & - & 97.58 & - & - & - \\
     $\textrm{GeoLayoutLM}$ & Linear/MLP & - & - & 92.86 & - & - & \bf 97.97 & - & - & - \\
     $\textrm{GraphLayoutLM}_{\rm LARGE}$ & Linear/MLP & 94.49 & 94.30 & 94.39 & 97.75 & \bf 97.75 & 97.75 & - & - & - \\
     \midrule 
     $\textrm{GraphLayoutLM}_{\rm LARGE}^\dagger$ & Linear/MLP & 94.37 & 93.95 & 94.16 & 97.32 & 97.68 & 97.50 & 99.27 & \bf 99.58 & 99.42 \\
     $\textrm{HGALayoutLM}_{\rm LARGE}$ & HGA & \bf 95.67 & \bf 94.95 & \bf 95.31 & \bf 97.97 & 97.38 & 97.67 & \bf 99.69 & 99.53 & \bf 99.61  \\
     \bottomrule
    \end{tabular}
    }
\end{table*}

\begin{table}[t]
    \centering
    \caption{Precision, Recall and F1 Score of Results on XFUND Datasets. Model labeled with "$^\dagger$" indicate that its results are obtained through replication in our experiments.}
    \label{tab:xfund_results}
    \scalebox{0.8}{
    \begin{tabular}{l|l|ccc}
    \toprule
    \multirow{2}{*}{\bf Model} & \multirow{2}{*}{\bf Head} &\multicolumn{3}{c}{\bf XFUND} \\
    & & \bf P & \bf R  & \bf F \\
     \midrule 
     $\textrm{LayoutXLM}_{\rm BASE}$ & Linear & - & - & 89.24 \\
     $\textrm{XYLayoutLM}$ & Linear & - & - & 91.76 \\
     $\textrm{LayoutLMv3}_{\rm BASE}$ & Linear & 89.80 & 94.35 & 92.02 \\
     $\textrm{GraphLayoutLM}_{\rm BASE}$ & Linear & 91.80 & 95.38 & 93.56 \\
     \midrule 
     $\textrm{GraphLayoutLM}_{\rm BASE}^\dagger$ & Linear & 92.30 & 94.69 & 93.48 \\
     $\textrm{HGALayoutLM}_{\rm BASE}$ & HGA & \bf 92.79 & \bf 95.70 & \bf 94.22 \\
     \bottomrule
    \end{tabular}
    }
\end{table}

\paragraph{Model Settings.}

The model settings are consistent with those of GraphLayoutLM.  The text sequence length is 512 and the document image is resized to $3 \times 224 \times 224$ dimensions.  The image is cut into 196 patches in the size of $16 \times 16$.  Transformer self-attention layer scaling factor $\alpha$ is set to 32.  For $\text{HGALayoutLM}_{\text{BASE}}$, the hidden layer dimensions, the number of encoder self-attention layers, the number of self-attention heads and intermediate dimensions for feed-forward networks are set to 768,12,12 and 3072, respectively.  The head number of graph mask layer is 6.  The hidden layer dimension, encoder self-attention layer number, self-attention head number and intermediate dimensions for feed-forward networks of $\text{HGALayoutLM}_{\text{LARGE}}$ are set to 1024,24,16 and 4096, respectively.  The head number of graph mask layer is 8.  The hidden size of hypergraph attention layer in both base and large model is set to 64. To ensure the fairness of the experiment, we convert the results of hypergraph extraction into the format of BIO annotations for comparison.

\paragraph{Datasets.}

We select four commonly used document information extraction datasets. Three of these datasets are in English, including FUNSD, CORD and SROIE. The other is the Chinese dataset, XFUND. The current XFUND task semantic entity recognition task of comparative experiment results is less and there is almost no LARGE version  experiment results. We only choose the BASE version of the model for our experiments. 
Detailed dataset information and finetuning hyper-parameters settings can be viewed in Tables \ref{tab:datasets_data} and \ref{tab:hyper_parameter}, respectively.

\paragraph{Baselines.}

We choose the classical natural language processing model BERT~\cite{devlin2018bert} as the single modal document understanding comparison model and select several classical multimodal document understanding models, such as LayoutLM~\cite{xu2020layoutlm}, BROS~\cite{hong2022bros}, LayoutLMv2~\cite{xu2020layoutlmv2} and LayoutXLM~\cite{xu2021layoutxlm}. We also include the latest works in document understanding for comparison, such as ERNIE-Layout~\cite{peng2022ernie}, LayoutLMv3~\cite{huang2022layoutlmv3}, GeoLayoutLM~\cite{luo2023geolayoutlm}, GraphLayoutLM~\cite{li2023enhancing} and UDop~\cite{tang2023unifying}. 
It is worth noting that according to the code design of LayoutLMv3 and GraphLayoutLM, different heads are selected under different conditions of the number of label types.
Specifically, the model uses linear layer as the classification head when there are less than 10 types of labels (e.g. FUNSD, SROIE, XFUND).
On the contrary, when the number of labels is greater than or equal to 10 (e.g. CORD), the MLP is selected as the classification head.

\subsection{Main Results}

\begin{figure}[t]
\centering

\begin{tikzpicture} [scale=0.90]
\begin{axis}[
    xlabel=step, 
    ylabel=F1, 
    legend pos=south east
    ]

\addplot[color=blue,mark=square,] table {src/position_encoding/wo_position.dat};
\addlegendentry{w/o pos}

\addplot[color=red,mark=*,] table {src/position_encoding/w_position.dat};
\addlegendentry{w/ pos}

\addplot[color=orange,mark=triangle,] table {src/position_encoding/w_span_position.dat};
\addlegendentry{w/ span pos}

\end{axis}
\end{tikzpicture}

\caption{Position Encoding Comparison Line Chart. In order to highlight the contrast effect, we omit the results for the first 300 steps when the model has not converged.}
\label{fig:position_encoding}
\end{figure}
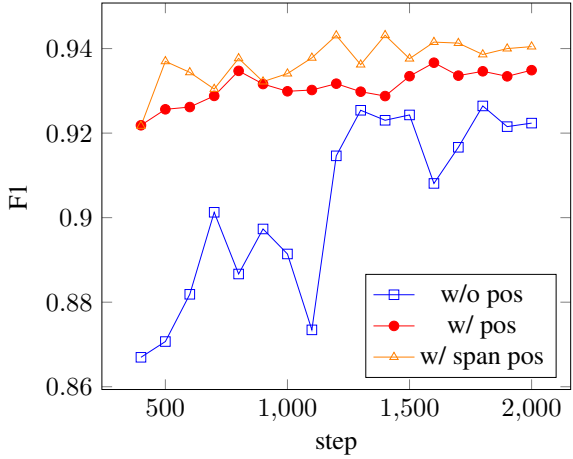

\begin{figure}[t]
\centering

\begin{tikzpicture} [scale=0.90]
\begin{axis}[
nodes near coords,
xlabel={b}, 
ylabel={F1}, 
enlargelimits=0.05,
ybar,
 ytick={97.0,97.2,97.4,97.6},
 yticklabels={97.0,97.2,97.4,97.6},
enlargelimits=true
]
\addplot[
    color=blue!30,
    fill=blue!30,
    bar width=20pt
] 
coordinates {
(0 ,96.91) 
(0.2 ,96.99)
(0.4 ,97.52) 
(0.6 ,97.04)
(0.8 ,97.11)};
\end{axis} 
\end{tikzpicture}
\caption{Further Study of Balance Factor.}
\label{fig:further_study_of_b}
\end{figure}
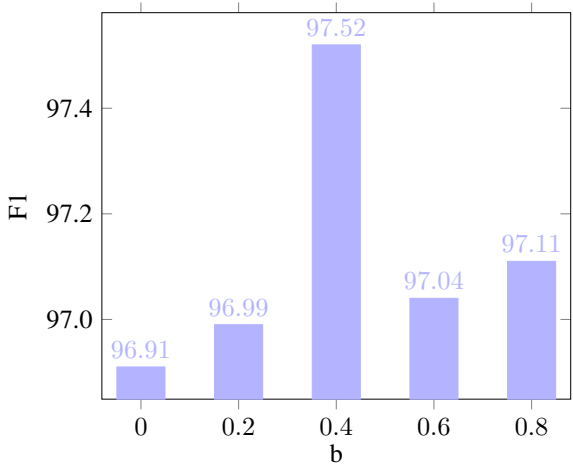

\begin{table}[t]
    \small
    \centering
    \caption{F1 Score of Different Head. }
    \label{tab:different_classifier}
    \scalebox{1}{
    \begin{tabular}{c|ccccc}
    \toprule
    \bf Head & \bf FUNSD & \bf CORD & \bf SROIE & \bf XFUND \\
    \midrule 
    Linear & 93.48&96.98&98.99&93.03 \\
    MLP & 93.58&97.13&99.28&93.48 \\
    HGA & 94.32&97.52&99.53&94.22 \\
    \bottomrule
    \end{tabular}
    }
\end{table}

\begin{table}[t]
    \caption{Analysis of Time and Space Complexity.}
    \label{tab:time_and_space_complexity}
    \centering
    \begin{tabular}{c|c|cc}
    \toprule
    \bf Model & \bf Head & \bf Params & \bf Flops \\ \midrule 
    GraphLayoutLM & Linear & 88.02M & 63.03G \\ 
    GraphLayoutLM & MLP & 88.61M & 63.45G \\ 
    HGALayoutLM & HGA & 88.31M & 63.24G \\ 
    \bottomrule
    \end{tabular}
\end{table}

The English datasets experiment results are shown in Table \ref{tab:main_results}. The BASE version of HGALayoutLM using hypergraph attention layer as the head has achieved the best results on FUNSD and SROIE datasets (94.32 on FUNSD and 99.53 on SROIE), even when compared to the LARGE versions of models. Compared with $\text{GraphLayoutLM}_{\text{BASE}}$ using linear classification, HGALayoutLM achieves improvements of 0.89, 0.39 and 0.54 on FUNSD, CORD and SROIE datasets, respectively. The LARGE version of HGALayoutLM has achieved F1 scores of 95.31 and 99.61 on FUNSD and SROIE respectively, further updating the best performance on these datasets. Compared with GraphLayoutLM in the LARGE version, HGALayoutLM has F1 score 1.15 and 0.19 higher on FUNSD and SROIE datasets, respectively. This demonstrates the effectiveness of HGA on the task of less labels.

However, we can find that the performance of HGA is not outstanding on the CORD dataset.  We think this is because the CORD dataset has a large number of label categories.   The number of labels in CORD is an amazing 30, compared with the 3 or 4 label categories in other datasets.   Since in the process of constructing the hypergraph, different types of hyperedges are built separately.   Plenty of label categories will make the effective span nodes of hypergraph matrix sparse, which is not conducive to semantic entity recognition.  However, by comparing GraphLayoutLM, we can find that HGA head can still improve the performance.

The experiment results of Chinese dataset, XFUND, are shown in Table  \ref{tab:xfund_results}. We can find that our HGALayoutLM has achieved the state of the art in XFUND (Precision 92.79, Recall 95.70 and F1 94.22). This further verifies the effectiveness of HGA head. 

\begin{table*}[t]
    \caption{Comparison with LayoutLLM. The results on FUNSD and CORD are F1.}
    \label{tab:comparison_with_layoutllm}
    \centering
    \begin{tabular}{c|cc|cc}
    \toprule
    \bf Model & \bf FUNSD & \bf CORD & \bf Params & \bf Flops \\ \midrule 
    $\textrm{HGALayoutLM}_{\rm LARGE}$ & 95.3 & 97.7 & 307.7M & 218.95G \\ 
    LayoutLLM & 95.3 & 98.6 & 6914.38M & 8654.62G \\ 
    \bottomrule
    \end{tabular}
\end{table*}

\subsection{Ablation Study}

To verify the effectiveness of our Span Position Encoding. We conduct ablation study on FUNSD. We can see from Figure \ref{fig:position_encoding} that the entity extraction effect without position encoding (w/o pos) is much worse than that with position encoding. In addition, we also compare the performance of our span position encoding (w/ span pos) with that of traditional position encoding (w/ pos). We can find that the performance of our span position encoding is obviously better than that of traditional position encoding.This demonstrates the effectiveness of our span position encoding with span prompt.

In order to prove that Balanced Hyperedge Loss can solve the problem of sparse hyperedge matrix caused by too many entity types. We conduct experiment statistics on different value of balance factor on CORD dataset with plenty of entity types and present the results in Figure \ref{fig:further_study_of_b}.  We can see that the performance of the unbalanced model ($b=0$) is not ideal, even worse than the performance of the MLP head. However, proper balance factor allow the model to pay more attention to the hyperedge entities and achieve better results. For example, the performance when $b$ is 0.4 exceeds the performance when the MLP is used as the head.

\subsection{Anaysis of Different Head}

To analyze the effects of different head, we adopt $\text{GraphLayoutLM}_{\text{BASE}}$ and $\text{HGALayoutLM}_{\text{BASE}}$ as the base model to conduct comparative experiments on three different heads: linear layer, MLP and HGA. The experiments are carried out on FUNSD, CORD, SROIE and XFUND datasets.

The experiment results are shown in Table \ref{tab:different_classifier}. As the simplest network structure, the linear layer has the worst classification effect. The MLP increases the number of linear layers on top of the linear layer. It also joins activation layers and dropout layers to linear layers. The more complex network structure makes MLP slightly better than the semantic entity recognition of a single linear layer on most datasets. As our proposed hypergraph attention method, HGA performs significantly better than the other two classifiers,which shows the effectiveness of HGA, which demonstrates the superior performance of HGA.

To test the complexity of HGA, we compare HGALayoutLM with the model with traditional heads. The PyTorch-OpCounter tool \footnote{https://github.com/Lyken17/pytorch-OpCounter} is used to calculate the time and space complexity.  The number of entity types is set to 3.  As we can see from Table \ref{tab:time_and_space_complexity},  HGA does not bring a large cost of time and space calculation  and HGA is even less costly than MLP in terms of time and space computation. This indicates that our performance improvement is not due to the increase in the number of parameters.

\subsection{Comparison with Large Language Model}

We conduct a comparative analysis of our HGALayoutLM with the latest document multimodal large language model, LayoutLLM~\cite{luo2024layoutllm}, to analyze the advantages and disadvantages of the models. LayoutLLM, which uses LayoutLMv3 as encoder and Llama as decoder, has so far achieved the state of the art results on several document intelligence tasks. We show the comparison results in Table \ref{tab:comparison_with_layoutllm}. We can see that our HGALayoutLM slightly underperforms compared to the professionally fine-tuned document large language model. However, under the premise of similar performance to large language models, our model parameters and computational consumption are much lower than the existing large language models. This fully demonstrates the advantage of our method.

\section{Conclusion}

In this work, we propose a semantic entity recognition method (HGA) based on hypergraph attention.  This method extracts semantic information from documents by establishing different hyperedges between feature nodes.  On the basis of the hypergraph, we design span position encoding and balanced hyperedge loss to enhance the entity extraction capability of the hypergraph attention head.  We use the HGA method to build a novel semantic entity recognition model HGALayoutLM based on GraphLayoutLM.  This model has good performance in SER tasks.  Experiments show that our method achieves the state of art on semantic entity recognition tasks on the FUNSD and XFUND datasets.

\section{Limitation}
The HGA method can achieve good performance on semantic entity recognition tasks, but there is still a lot of work for us to improve.
On the one hand, when there are more types of semantic entities, the cost of improvement from HGA becomes higher. The number of superedge matrices increases because of more semantic entity categories. This not only leads to sparse label matrices, but also to more model parameters. How to solve the matrix sparsity and parameter growth caused by the number of label types is the future work we need to study. On the other hand, since our proposed head is currently targeted at semantic entity recognition tasks in the document domain. In the future, we will explore more general head to adapt to diverse document task types.

\newpage
\bibliography{anthology,custom}

\end{document}